\documentclass{hld2025} 

\usepackage{amsmath}
\usepackage{amssymb}
\usepackage{mathtools}
\usepackage{booktabs}
\usepackage{placeins}

\usepackage[capitalize,noabbrev]{cleveref}

\usepackage[textsize=tiny]{todonotes}

\usepackage{bm}
\usepackage{braket}
\usepackage[T1]{fontenc}

\newcommand{\vx}{\bm{x}}
\newcommand{\vy}{\bm{y}}
\newcommand{\vX}{\bm{X}}
\newcommand{\vY}{\bm{Y}}
\newcommand{\vtheta}{\bm{\theta}}
\newcommand{\vh}{\bm{h}}
\newcommand{\vb}{\bm{b}}
\newcommand{\vW}{\bm{W}}
\newcommand{\vv}{\bm{v}}
\newcommand{\eg}{\mathcal{E}_\textrm{gen}}

\definecolor{blue1}{RGB}{204,229,255}
\definecolor{blue2}{RGB}{0,76,153}

\usepackage{pifont}

\usepackage{tcolorbox}
\tcbuselibrary{skins, breakable}
\tcbset{
    boxrule=0pt,
    colframe=gray!90,
    colback=gray!20,
    coltitle=white,
    enhanced,
    boxsep=5pt,
    arc=2mm}

\title[Disentangling Feature Learning from Generalization in Neural Networks]{Disentangling Feature Learning from Generalization in Neural Networks}

\hldauthor{%
\Name{Niclas G{\"o}ring} \Email{niclas.goring@physics.ox.ac.uk}\\
\addr {University of Oxford, Oxford, UK} 
\AND
\Name{Charles London} \Email{charles.london@cs.ox.ac.uk}\\
\addr {University of Oxford, Oxford, UK} 
\AND
\Name{Hadi Ertuk} \Email{abdurrahman.erturk@physics.ox.ac.uk}\\
\addr {University of Oxford, Oxford, UK} 
\AND
\Name{Chris Mingard} \Email{christopher.mingard@queens.ox.ac.uk}\\
\addr {University of Oxford, Oxford, UK} 
\AND
\Name{Yoonsoo Nam} \Email{yoonsoo.nam@physics.ox.ac.uk}\\
\addr {University of Oxford, Oxford, UK} 
\AND
\Name{Ard A. Louis} \Email{ard.louis@physics.ox.ac.uk}\\
\addr {University of Oxford, Oxford, UK} 
}

\begin{document}

\maketitle

\begin{abstract}%
Neural networks outperform kernel methods, sometimes by orders of magnitude, e.g. on staircase functions. This advantage stems from the ability of neural networks to learn features, adapting their hidden representations to better capture the data. We introduce a concept we call feature quality to measure this performance improvement. We examine existing theories of feature learning and demonstrate empirically that they primarily assess the strength of feature learning, rather than the quality of the learned features themselves. Consequently, current theories of feature learning do not provide a sufficient foundation for developing theories of neural network generalization. 

\end{abstract}

\section{Introduction}

Neural networks (NNs) generalize remarkably well in diverse domains, from computer vision to natural language processing or to protein folding \cite{brown2020language,lecun2015deep,jumper2021highly}. However, understanding the mechanisms behind this success remains a fundamental challenge in machine learning. A leading hypothesis attributes this success to feature learning (FL) -- the network's ability to adapt its hidden representations to discover useful patterns from data. For instance, visualization techniques reveal that convolutional NNs naturally develop hierarchical feature representations, progressively learning to detect edges, textures, patterns, object parts, and finally complete objects \cite{olah2017feature}. When comparing NNs to their linearized approximations \cite{Jacot_Gabriel_Hongler_2018}, NNs achieve dramatically better sample complexity on many tasks (see discussion in \cref{subsec_FLgap}). This suggests that NNs learn better features through training than those present at initialization. Current literature characterizes FL as a change in the Neural Tangent Kernel (NTK, \cite{Jacot_Gabriel_Hongler_2018}), Conjugate Kernel (CK, \cite{Lee_Bahri_Novak_Schoenholz_Pennington_Sohl-Dickstein_2018}), or via some weight-based metric. These approaches measure FL by quantifying how much a trained network deviates from its linear approximation at initialization. While there is a general notion that FL improves generalization, we argue this relationship is misleading: FL theories measure FL strength -- the magnitude of representation change -- which is fundamentally decoupled from feature quality -- the actual impact on generalization. Our \textbf{contributions} are: \emph{(i)} We define a rigorous way to measure feature quality through the FL gap $\Delta_{\operatorname{NT}}$. \emph{(ii)}  We demonstrate that current FL definitions measure strength rather than quality, and show these are decoupled. 
See \cref{appx_background} for notation and background on kernel methods and \cref{appx_relwork} for related work.

\begin{figure*}[htb]
\vskip 0.2in
\begin{center}
\centerline{\includegraphics[width=12.9cm]{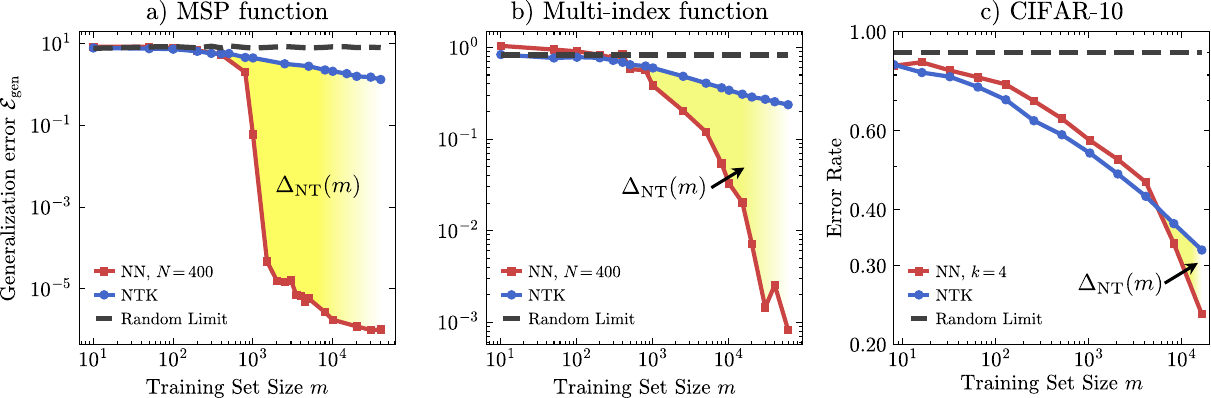}}
\caption{\textbf{Generalization error} $\eg$ \textbf{versus training set size} $m$ for NNs and their corresponding NTK across three distinct target functions: (a) FFNN on merged staircase (MSP) functions  \cref{appx_sec_msp_f}, (b) FFNN on Multi-index functions \cref{appx_sec_pp_f}, and (c) Wide ResNet on CIFAR-10. For (a) and (b), we observe a critical training set size $m^*$ where NNs outperfrom their NTK counterparts by orders of magnitude ($m^*\sim 10^3$). We quantify this improvement in performance through the FL gap $\Delta_{\operatorname{NT}}$ (\cref{def_FLgap}).
For (c), the learning curve for the NN scales similarly to the NTK until $\sim 10^4$. 
}
\label{fig_fig1}
\end{center}
\vskip -0.2in
\end{figure*}


\section{Feature Learning vs. Feature Quality}
\label{sec_flgap}

\subsection{The feature learning gap}
\label{subsec_FLgap}
\cref{fig_fig1} shows that the NTK significantly underperforms the corresponding NN after some amount of data $m^*\sim10^3$ on the MSP functions and multi-index functions. On CIFAR-10, the NTK achieves comparable performance to the NN until $m^*\sim10^4$.\footnote{This is well predicted by how well the empirical NTK aligns with the target function see Appx. \cref{fig_cumm}.} MSP and multi-index functions are not isolated examples; rather, there exists a large class of target functions (collected in \cref{tab:nn_kernel_comparison}) where kernels have polynomial or even exponential sample complexity, and NNs achieve linear or lower degree polynomial sample complexity. These dramatic differences in sample complexity suggest that the NNs learn high-quality features not present at initialization (and thus in their CK/NTK). We quantify this feature quality through the FL gap.

\begin{definition}[Feature learning gap]\label{def_FLgap} 
Given a data generating model $p(\vx,\vy)$, an i.i.d. dataset $\mathcal{D}$ of size $m$ with a target function $f^*:\mathbb{R}^{n_0} \rightarrow \mathbb{R}^{n_{L}}$, a FFNN $f_{\vtheta}$, and the mean predictor of the FFNN's NTK, given by $\mu_{\operatorname{NT}}(\vx) = K_{\operatorname{NT}}(\vx,\vX)N_{\operatorname{NT}}(\vX,\vX)^{-1}\vY$, the FL gap is defined by
  \begin{align}
  \Delta_{\operatorname{NT}}(m)=\mathcal{E}_{\operatorname{gen}}(\mu_{\operatorname{NT}};m) -\mathcal{E}_{\operatorname{gen}}(f_{\vtheta};m).
  \end{align}
\end{definition}
See \cref{appx_sec_conc} for further a discussion on properties of $\Delta_{\operatorname{NT}}$.
\subsection{Disentangling FL strength from feature quality} \label{sec:strength_quality}
The literature contains multiple definitions of FL, falling into three main categories: (1) NTK-based, (2) CK-based, and (3) superposition-based definitions. These approaches characterize FL by measuring how hidden representations of $f_{\vtheta(t)}$ change during training relative to the initialized network $f_{\vtheta(0)}$. Although these measures take different forms based on changes in the NTK, CK, or other metrics, we argue they fundamentally measure \textit{FL strength} rather than \textit{feature quality} (i.e. how useful the features are). This is based on our empirical observations in \cref{sec_current_fl_def} indicating 
that changes in hidden representations do not guarantee the learning of high-quality features, as quantified by $\Delta_{\operatorname{NT}}$. Conversely, an NN can generalize effectively with minimal changes to its representations if the initial kernel already encodes useful features \cite{petrini_learning_2022}.

\begin{tcolorbox}[colback=gray!5, colframe=black, boxsep=0pt, left=6pt, right=6pt, top=6pt, bottom=6pt]
 \textbf{Claim:} Current FL definitions (explicitly or implicitly) characterize FL by measuring FL strength $S(f_{\vtheta})$. However, FL strength is decoupled from feature quality, measured by the FL gap $\Delta_{\operatorname{NT}}$.
\end{tcolorbox}

\section{Current FL definitions do not provide a feature quality measure} 
\label{sec_current_fl_def}

\paragraph{Methodology} Every FL theory provides a measure of feature strength $S(f_{\vtheta})$. Our goal in this section is to assess whether the strength measures correlate with the FL gap $\Delta_{\operatorname{NT}}$, i.e.\ whether strong FL implies a beyond-the-kernel performance. 
We conduct experiments with two architectures and datasets (1) CNNs trained on CIFAR-10 and (2) FFNNs trained on MSP functions \footnote{MSP functions map to whole numbers, making label shuffling well-defined, see \cref{appx_sec_msp_f}.}. For each setup, we compare models trained on true-labeled data to those trained on shuffled labels. Any non-vacuous generalization bound must be data-dependent \cite{Arpit2017Memorization,Zhang_Bengio_Hardt_Recht_Vinyals_2017}. If $S(f_{\vtheta})$ correlates with feature quality ($\Delta_{\operatorname{NT}}$), the feature strength measures must demonstrate a qualitative distinction between NNs trained on shuffled vs. non-shuffled data. Otherwise, the result suggests a lack of correlation between feature strength and quality. 

\subsection{Family 1: NTK based definitions}

The first FL definition we examine is based on the identification of two training regimes, the ``lazy'' and ``rich'' regimes \cite{Moroshko_Woodworth_Gunasekar_Lee_Srebro_Soudry_2020,chizat_lazy_2020}. In the lazy regime, NNs behave like their linearized approximations
\begin{equation}
\label{eq_linear}
    f_{\vtheta}(\vx) = f_{\vtheta_0}(\vx) + (\vtheta - \vtheta_0)^\top \nabla_{\vtheta} f_{\vtheta_0}(\vx)+\mathcal{O}(\vtheta^2).
\end{equation}
Although multiple FL definitions exist in the literature \cite{karkada_lazy_2024,domine_lazy_2024,Tu_Aranguri_Jacot_2024,Seleznova,Moroshko_Woodworth_Gunasekar_Lee_Srebro_Soudry_2020,Wang_Wu_Lee_Ma_Ge_2020,Geiger_Spigler_Jacot_Wyart_2020}, they are fundamentally related. NNs FL when they diverge from their linearization at initialization eq. \eqref{eq_linear}. This can be measured by the change in the NTK.

\begin{definition}[Feature learning (NTK)]\label{def_lazy}
    A NN $f_{\vtheta}$ feature learns if the empirical NTK $\hat{K}_{\operatorname{NT}}$ changes significantly during training $ \exists \varepsilon, T > 0: \  \forall t >T \quad d( \hat{K}_{\operatorname{NT}}(\vtheta_0) ,  \hat{K}_{\operatorname{NT}}(\vtheta_t)) > \varepsilon $, 
    where $d$ is some distance metric for kernels.
\end{definition}
\begin{definition}[Feature]
The features are the row vectors of the feature map: $\Phi(\vx;t)= \nabla_{\vtheta} f_{\vtheta}(\vx)|_{\vtheta_t}$.
\end{definition}
For NTK-based FL definitions, a larger distance between the initial and final NTK correlates with stronger FL. Accordingly, the following definition of FL strength seems natural.

\begin{definition}[FL strength (NTK)]
\label{def_fquality1}
We set $S_{\operatorname{NT}}(f_{\vtheta})=1-\kappa_{\operatorname{CKA}}( \hat{K}_{\operatorname{NT}}(\vtheta_0),  \hat{K}_{\operatorname{NT}}(\vtheta_t)),
    $ where $\kappa_{\operatorname{CKA}}$ is the centered-kernel alignment \cite{Kornblith_Norouzi_Lee_Hinton_2019} which measures the normalized distance between kernels. 
\end{definition}
\begin{figure*}[htb!]
\begin{center}
\centerline{\includegraphics[width=12.9cm]{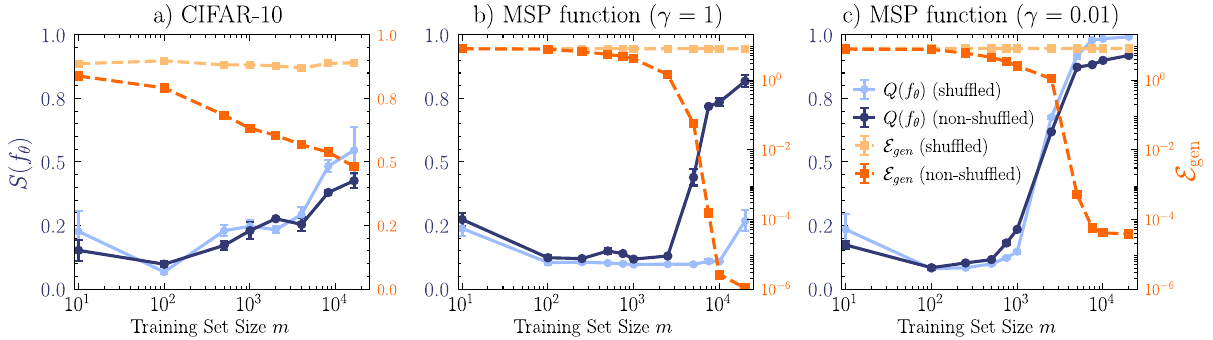}}
\vskip -0.1in
\caption{
\textbf{FL strength $S_{\operatorname{NT}}(f_{\vtheta})$ is decoupled from generalization error $\eg$.} 
(a) shows a CNN on CIFAR-10 and (b,c) an FFNN on MSP functions with true and shuffled labels.
(b) clearly shows significant difference in $S_{\operatorname{NT}}(f_{\vtheta})$ between the NN and corresponding NTK after $m^*\sim10^3$, 
However, this difference vanishes when scaling the network output by $\gamma=0.01$, shown in (c), with no corresponding change in $\eg$. This indicates $S_{\operatorname{NT}}(f_{\vtheta})$ is not predictive of $\eg$. As in \cref{fig_fig1} there is no significant difference between the NN and corresponding NTK for the CNN in (a).
See \cref{tab_fig2_CNN,tab_fig2_FFNN} for the architecture.
}
\label{fig_cifar_shuffle}
\end{center}
\vskip -0.2in
\end{figure*}

\paragraph{Critique}
 $S(f_{\vtheta})$ is calculated using the neural-tangents package \cite{neuraltangents2020,novak2022fast} as per \cref{def_fquality1}. For a CNN trained on CIFAR-10, $S(f_{\vtheta})$ fails to distinguish between shuffled and non-shuffled labels (\cref{fig_cifar_shuffle}(a), \cref{tab_fig2_CNN}). While experiments with an FFNN on MSP functions initially seem promising—with $S_{\operatorname{NT}}(f_{\vtheta})$ increasing for non-shuffled data around $m=3000$ unlike the shuffled case (\cref{fig_cifar_shuffle}(b), \cref{tab_fig2_FFNN})—this signal is not robust. This apparent difference can be nullified by introducing a scaling parameter $\gamma$ \cite{chizat_lazy_2020,Atanasov_Meterez_Simon_Pehlevan_2024}, where the network output becomes $\tilde{f}_{\vtheta}(\vx) = \frac{1}{\gamma}f_{\vtheta}(\vx)$. As shown in \cref{fig_cifar_shuffle}(c), setting $\gamma=0.01$ makes both the FL strength and the learning curves qualitatively indistinguishable for shuffled and non-shuffled data. Because the generalization error is largely unaffected by this scaling, the metric's sensitivity to $\gamma$ demonstrates that $S_{\operatorname{NT}}(f_{\vtheta})$ is not a robust predictor of generalization. This aligns with recent findings on "misgrokking" \cite{Lyu_Jin_Li_Du_Lee_Hu_2024}, where NTK changes can decouple from generalization.

\subsection{Family 2: CK based definitions}
A second line of work bases features on the CK.  While  \citet{Yang_Hu_2021,Nam_Mingard_Lee_Hayou_Louis_2024} are focused on the final layer CK, \citet{Naveh_Ringel_2021,Seroussi_Naveh_Ringel_2023,Fischer_Lindner_Dahmen_Ringel_Krämer_Helias_2024} treat CKs of each layer in a Bayesian framework.
\begin{definition}[Feature Learning]
An NN undergoes FL if its final layer feature map $\Phi^{L-1}(\vx;t)$ differs from its initialization $\Phi^{L-1}(\vx;0)$ at any time $t$ for some input $\vx \in \mathcal{X}$.
\end{definition}
\begin{definition}[Feature]
Given an NN, the features are the eigenfunctions $[e_1(t),...,e_{n_L}(t)]$ of the last layer CK, $ K_{\Phi^{L-1}}(\vx_{\mu},\vx_{\nu};t)$ (see \cref{subsec_kernelmethods}), ordered by their eigenvalues $\lambda_k$.
\end{definition}

\begin{definition}[FL strength (CK)]\label{def_yoonsoo_quality}
Given the trained (scalar-valued) network $f_{\vtheta}$, the utility of the $k$-th feature $e_k(t)$ is $\hat{Q}_k=\braket{e_k|f_{\vtheta}}^2$. The cumulative utility  of the first $k$ features is $\hat{\Pi}(k) = \sum_{j=1}^k \hat{Q}_j$, with $0 \leq \hat{\Pi}(k) \leq 1$ and $\hat{\Pi}(n_L) = 1$. We define the FL strength (CK) as
$   S_{\operatorname{CK}}(f_{\vtheta})=\min_k\{ \hat{\Pi}(k)> \varepsilon  \},$
where $\varepsilon=0.95$ is chosen as a sensible threshold.
\end{definition}
When $\hat{\Pi}(k)$ approaches $1$ quickly with $k$, it means that the NN is strongly  learning features, akin to neural collapse \cite{Papyan_Han_Donoho_2020} where only a minimal number of features are used.

\begin{figure}[htb!]
\begin{center}
\centerline{\includegraphics[width=14.0cm]{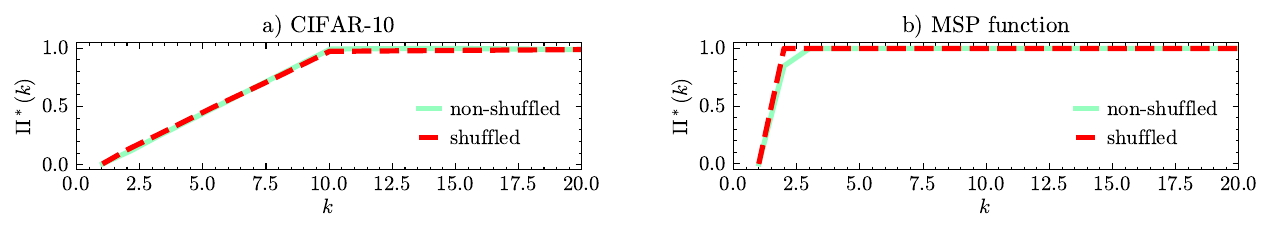}}
\vskip -0.1in
\caption{\textbf{The Cumulative quality of features $\Pi^*$ is decoupled from generalization.} (a) a ResNet on CIFAR-10 and (b) an FFNN trained on MSP functions, each with shuffled and non-shuffled data.
}
\label{fig_yoonsoo_quality}
\end{center}
\vskip -0.2in
\end{figure}

\paragraph{Critique}
The cumulative utility metric fails to distinguish between networks trained on true versus randomly shuffled labels. This holds for both CIFAR-10 (\cref{fig_yoonsoo_quality} (a)) and for FFNNs trained on the MSP function (\cref{fig_yoonsoo_quality} (b)). Our analysis of the CK spectra also reveals qualitatively similar patterns for both shuffled and non-shuffled data (\cref{appx_sec_2family_spectra,appx_cksp1,appx_cksp4}). These findings align with previous research showing that neural collapse, which is equivalent to strong feature learning under CK-based definitions, can occur independently of generalization \cite{Kothapalli_2023,Hui_Belkin_Nakkiran_2022,Galanti_Galanti_Ben-Shaul_2022}.


\subsection{Family 3: Superposition based definitions}
\citet{elhage_toy_2022} define features as ``\textit{properties of the input which a sufficiently large NN will reliably dedicate a neuron to representing}''. While this definition needs further elaboration as we do not know when an NN is ``sufficiently large'', they later give a more practical definition of a feature which is closely related to family 2.

\begin{definition}[Feature]
Given a FFNN with a feature map $\Phi^k(t)$ as defined in \cref{def_featuremap}, a feature $f_i$ corresponds to a direction $\vv_i \in \mathbb{R}^{n_k}$ in the hidden (activation) space.
\end{definition}
Features correspond to hidden-space vectors $\vv_i $, whose count can exceed the layer width $n_k$. This non-orthogonal “superposition” allows more features than dimensions \cite{arora_linear_2018,hanni_mathematical_2024}. Given features with values $x_{f_1},x_{f_2},\ldots$, the layer encodes them as $\Phi^k(\vx_{\mu};t)\;=\;\sum_{i=1}^{n_k} x_{f_i}\,\vv_i $. \cite{elhage_toy_2022} quantified superposition in autoencoders via \emph{feature} and \emph{sample} dimensionality. There is no canonical way to generalize these measures from an autoencoder architecture to a FFNN in the overparameterized regime. Nevertheless, here we adopt a layer-wise definition:


\begin{definition}[FL strength]
For layer $k$, FL strength is measured through two complementary metrics
\vspace{-0.4em}
\begin{equation}
\begin{aligned}
D_{f_i} = \frac{||\vW_i||_2^2}{\sum_j (\hat{\vW}_i \cdot \vW_j)^2}, 
\quad \quad D_{\vx_{\mu}} = \frac{||\Phi^k(\vx_{\mu};t)||_2^2}{\sum_{\nu} (\hat{\Phi}^k(\vx_{\mu};t) \cdot \Phi^k(\vx_{\nu};t))^2},
\end{aligned}
\end{equation}
feature dimensionality $D_{f_i}$ for feature $f_i$ and sample dimensionality $D_{\vx_{\mu}}$ for input $\vx_{\mu}$, where ``$\ \hat{} \ $`` denotes normalized vectors.
\end{definition}
Feature dimensionality measures how much of a hidden dimension is `dedicated' to representing a specific feature, ranging from 0 to 1. A feature with dimensionality 1 has its own dedicated dimension, while a feature with dimensionality closer to 0 is either not learned at all or shares its representation space with other features, existing in superposition. The same applies for sample dimensionality, but in terms of $\Phi^k(\vx_{\mu})$ rather than $\vW_i$.

\begin{figure}[htb!]
\vskip 0.2in
\begin{center}
\centerline{\includegraphics[width=14.0cm]{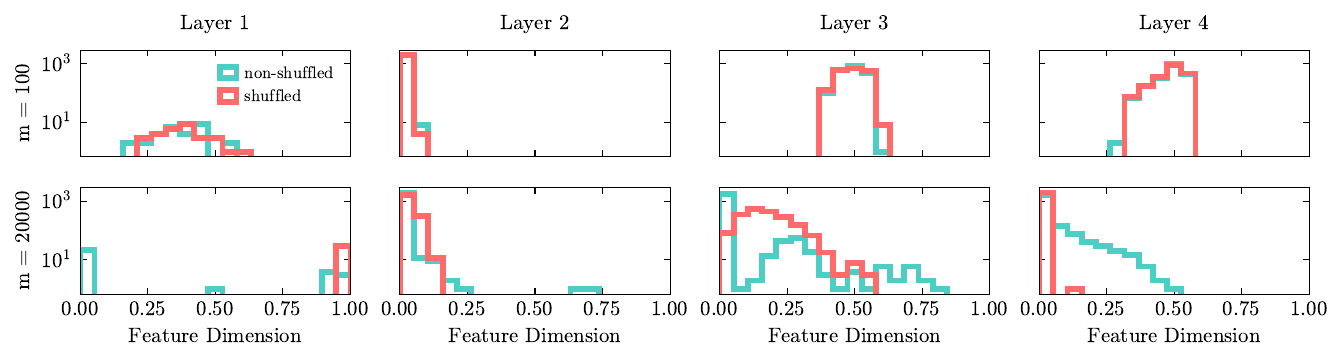}}
\caption{
\textbf{Feature dimensionality and generalization are not strongly correlated}
Histogram of the feature dimensionality for an FFNN with depth $L=4$ and width $N=2000$ trained on MSP functions for training set sizes $m=100$ and $20000$.
When $m=100$, there is no significant difference in the histograms, despite higher test losses for shuffled data (9.5 vs 6.3). At $m=20000$, shuffled data exhibits mostly zero $D_{f_i}$ with few non-zero features, a pattern consistent with FL, while non-shuffled data shows a diffuse distribution. The stark difference in test losses ($6.4$ vs $2\times 10^{-7}$) despite these patterns demonstrates that final layer feature dimensionality poorly predicts generalization.
}
\label{fig_anthro_main}
\end{center}
\vskip -0.2in
\end{figure}


\paragraph{Critique}
We trained an FFNN on MSP functions with shuffled and non-shuffled labels. While the sample dimensionality metric fails to distinguish between them, feature dimensionality reveals more nuanced patterns (see appx. \cref{fig_anthrohistg1}). The histograms in \cref{fig_anthro_main} for $m=100$ and $m=20000$ illustrate this. Specifically for $m=20000$, when enough data was available to learn features, the first layer's histogram for non-shuffled data shows an optimal pattern: most dimensions are near zero, with a few non-zero ones for important features. This pattern does not emerge in the data-poor case ($m=100$), is absent entirely in shuffled data, and dissolves in deeper layers. However, feature dimensionality is not a definitive measure of feature quality, as one generally does not know the relevant input features to validate the histogram. Furthermore, the distribution of $D_{f_i}$ can be heavily influenced by training parameters like $\gamma$ (appx. \cref{fig_anthrohistg00001}). Therefore, while $D_{f_i}$ is useful for measuring superposition, it is not a conclusive metric for feature quality.

\section{Conclusion}
\label{sec_conclusion}
We have demonstrated that current theories of FL, while capturing the strength of representation changes during training, mostly fail to predict generalization. This decoupling between FL strength and feature quality suggests the need for a more comprehensive FL theory if FL should act as a foundation for theories of NN generalization.


\bibliography{sample}

\newpage
\clearpage

\appendix
\newpage
\appendix
\onecolumn

\FloatBarrier

\section{Background}
\label{appx_background}
Let $\mathcal{X}\subseteq \mathbb{R}^{n_0}$ and $\mathcal{Y}\subseteq \mathbb{R}^{n_{L}}$ be the input and output space. A dataset of size $m$, $\mathcal{D}={(\vx_{\mu},\vy_{\mu})}_{\mu=1}^m$ is drawn i.i.d.\ from the data generating distribution $p(\vx,\vy)$.
For some function $f$, the generalization is defined with respect to a loss function $\ell:\mathcal{Y}\times \mathcal{Y} \rightarrow \mathbb{R}_{\geq 0}$,
\begin{equation}
\mathcal{E}_{\operatorname{gen}}(f) = \mathbb{E}_{(\vx,\vy)\sim p(\vx,\vy)}\left[\ell(f(\vx),\vy)\right].
\end{equation}
In practice, we will approximate this quantity by averaging over a finite test set.

\begin{definition}[Feed-forward Neural Network (FFNN)]
An $L$-layer FFNN is a recursively defined map $f_{\vtheta}:\mathbb{R}^{n_0} \rightarrow \mathbb{R}^{n_{L}} $:
\begin{equation}
\begin{aligned}
\vh^0 = \vx_\mu, \quad \vh^l &= \vW^l\phi(\vh^{l-1}) + \vb^l,
\end{aligned}
\end{equation}
and $f(\vx_\mu) = \vW^{L}\vh^{L-1}(\vx_{\mu}) + \vb^{L}$, where $1 \leq l\leq L$, $\vW^l \in \mathbb{R}^{n_l \times n_{l-1}}, \vb^l \in \mathbb{R}^{n_l}$, and $\phi$ are nonlinear functions applied element-wise. We assume all $n_l$ are equal for $l\neq 0, L$, and call this the width of the FFNN and denote $P$ for the total number of parameters.
\end{definition}

\subsection{Kernel methods}
\label{subsec_kernelmethods}
\begin{definition}[Kernel \& Features]\label{para_kernel}
A kernel is any symmetric, positive semi-definite function $K:\mathcal{X} \times \mathcal{X} \rightarrow \mathbb{R}$. 
Let $\mathcal{H}$ be the reproducing kernel Hilbert space (RKHS) with inner product $\braket{\cdot|\cdot}_{\mathcal{H}}$. Then, any such kernel can be written as an inner product kernel $K(\vx,\vx')=\braket{\Phi_K(\vx)|\Phi_K(\vx')}_{\mathcal{H}}$. The kernel's feature map is given by $\Phi_K: \mathcal{X} \rightarrow \mathcal{H}$.

\end{definition}

\begin{definition}[$l$-layer feature map]\label{def_featuremap}
    Consider the feature map of the $l<L$'th layer of an FFNN at training time $t$,
    \begin{equation}
        \Phi^l(t): \mathcal{X} \rightarrow \mathbb{R}^{n_l}, \quad \vx_{\mu} \mapsto \vh^l(\vx_{\mu}).
    \end{equation}
    The $l$'th layer feature kernel $K_{\Phi^l}(t): \mathcal{X}^2 \rightarrow \mathbb{R}$ is given by
    \begin{align}
        K_{\Phi^l}(\vx_{\mu},\vx_{\nu};t)=  \Phi^l(\vx_{\mu};t)^{\top} \Phi^l(\vx_{\nu};t).
    \end{align}
\end{definition}
When evaluated over a finite dataset $\mathcal{D}$, ${(K_{\Phi^l})}_{\mu\nu}$ can be interpreted as the correlation matrix, measuring how similar the features of $\vx_\mu$ and $\vx_\nu$ are at layer $l$.

\subsection{Learning dynamics and spectral bias for kernels}\label{sec:kernel_gen_exp}
When performing kernel ridge regression with gradient descent, the residual dynamics $r_t(\vx) = f(\vx) - f^*(\vx)$ for projections on eigenfunctions $e_{\rho}$ follow $\langle r_t| e_{\rho} \rangle_{\mathcal{H}} = e^{-\lambda_{\rho} t}\langle r_0| e_{\rho} \rangle_{\mathcal{H}}  $. For high-dimensional kernels, where the number of eigenfunctions $N_{\rho}$ greatly exceeds the number of samples $m$, the distribution over $\lambda_\rho$ determines the solution: eigenfunctions with large $\lambda_\rho$ will have large coefficients \cite{Geifman_Galun_Jacobs_Basri_2022,Geifman_Barzilai_Basri_Galun_2024,Rahaman_Baratin_Arpit_Draxler_Lin_Hamprecht_Bengio_Courville_2019}. Eigenfunctions with large $\lambda_{\rho}$ are learned the fastest, so a trained solution will be dominated by the corresponding eigenfunctions. These often correspond to low frequency, simple components of the target function, rigorously proven for ReLU networks in \cite{Basri_Galun_Geifman_Jacobs_Kasten_Kritchman_2020}. The high-frequency components, which could lead to overfitting, are naturally learned more slowly.

The generalization error $\eg$ scales as :
\begin{align}
   \mathcal{E}_{\operatorname{gen}}(m) \sim m^{-\beta}, \ \text{with} \ \beta = \frac{1}{d}\min(\alpha_T - d, 2\alpha_S)
\end{align}
where the exponent $\beta$ reflects how quickly different frequency components are learned, and $\alpha_T,\alpha_S$ are the decay rates of the kernel in Fourier space \citep{Spigler_Geiger_Wyart_2020,Bordelon_Canatar_Pehlevan_2020}. For $\beta$ to remain non-vanishing as dimension $d$ increases, the smoothness index $s = (\alpha_T-d)/2$ must scale with $d$ (curse of dimensionality).

\subsection{Kernels on CIFAR-10}
\label{appx_cifar} 
NTKs of CNNs are multi-dot product kernels $k(x,z)$ that operate over the multi-sphere $\Pi_{i=1}^d \mathbb{S}^{\zeta-1}$ where $d$ is the number of pixels and $\zeta$ is the number of channels \cite{Geifman_Galun_Jacobs_Basri_2022}. These kernels can be decomposed into eigenfunctions, which are multivariate spherical harmonics. The eigenvalue $\lambda_{\bm{k}}$ for the frequencies $\bm{k}$ of the multivariate spherical harmonic exhibits polynomial decay with respect to these frequencies. This decay induces an implicit bias that favors learning low-frequency functions before high-frequency ones, manifesting as a form of simplicity bias. \\ 
The multiplicity of the eigenvalues is determined by the quantity $p_i^{(L)}$, which represents the number of paths for a pixel $i$ in a network of depth $L$. This path count quantifies the distinct number of ways information from a pixel can propagate through the network's convolution layers to reach a particular output. For a pixel, the number of paths decays exponentially with the distance from the center of the receptive field, introducing a positional bias that facilitates learning spatially localized features over those requiring global image dependencies. This bias aligns with natural image statistics, where meaningful features typically exhibit local coherence. Consequently, CNNs can more efficiently learn localized high-frequency patterns compared to patterns requiring high frequencies across multiple pixels, a distinction not present in fully connected networks. This theoretical framework is further supported by \cite{Ghorbani_Mei_Misiakiewicz_Montanari_2020}, who demonstrate that such kernels generalize effectively when image labels depend on low frequencies (frequency bias) and the image spectrum itself is concentrated in low frequencies (positional bias), conditions commonly satisfied in real-world image datasets.\\
To conclude, alignment of the kernel with the target function can largely explain generalization of the NTK on specific datasets where this alignment exists, such as image classification, whereas when this alignment is absent, as in merged staircase functions, the NTK fails to generalize effectively.

\subsection{Generalization theory of Kernels}
To analyze the generalization behavior of the NTK, we need to first examine the theoretical foundations of generalization in kernel methods.  Given that the kernels eigenfunctions are a basis of the RKHS \ref{para_kernel}, we can decompose the predicted $ f^*(x) = \sum_\rho w^*_\rho \sqrt{\eta_{\rho}} \phi_\rho(x)$ and target function $\Bar{f}(x) = \sum_\rho \Bar{w}_\rho \sqrt{\eta_{\rho}} \phi_\rho(x)$ in terms of the eigenfunctions. This allows for decomposing the generalization error in terms of modes\footnote{There is a fundamental lower bound for the test error due to zero modes.}
\begin{equation}
\mathcal{E}(m) = \sum_{\rho} \eta_{\rho} \left\langle(w_\rho^* - \Bar{w}_\rho)^2\right\rangle_{\mathcal{D}} =\sum_{\rho}  \eta_{\rho} \mathcal{E}_{\rho}(m)
\end{equation}
where the average over datasets can be analytically computed \cite{Bordelon_Canatar_Pehlevan_2020}. We note that the spectrum $\{\eta_{\rho}\}$ is independent of the target function, while the mode error $\mathcal{E}_{\rho}$ is not.\\  As we focus on learning curves, we want to  understand how $\mathcal{E}(m)$ scales with $m$. Qualitatively, the scaling is dominated by two quantities \cite{Canatar_Bordelon_Pehlevan_2021,Bordelon_Canatar_Pehlevan_2020}. The first one is \textit{spectral alignment}. It can be formally shown that for $\eta_\rho > \eta_{\rho'} $, $ \mathcal{E}_\rho(m)$ decreases faster with $m$ than $ \mathcal{E}_{\rho'}(m)$. This means that, with growing training set size,  eigenfunctions with larger eigenvalues of the trained function approach the one of the target function faster. Hence, if the target function is well approximated by the high eigenvalue eigenfunctions of the kernel, the generalization error will drop faster. Secondly, the \textit{asymptotic mode error} has the form $\mathcal{E}_\rho(m) \underset{m \rightarrow \infty}{\sim} \frac{\braket{\Bar{w}_{\rho}}}{\eta_{\rho}}$. The asymptotic error of a mode is larger if the RKHS eigenvalue $\eta_{\rho}$ is small, even if the coefficient $\overline{w}_{\rho}$ of the target eigenfunction is large. Both of these observations motivate the definition of the cumulative power distribution. 
\begin{definition} \label{cumulative_power_distribution}
    The cumulative power distribution is defined as the amount of overlap of the target function with the RKHS subspace up to mode $\rho$:
    \begin{equation}
    C(\rho) = \frac{\sum_{\rho' \leq \rho}\eta_{\rho'}\overline{w_{\rho'}}^2}{\sum_{\rho'}\eta_{\rho'}\overline{w_{\rho'}}^2}
\end{equation}
\end{definition}
To conclude, the more power the target function has in the high eigenvalue subspace of the RKHS, the faster kernel ridge regression is able to learn the function with growing data set size.  High task-model alignment results in a faster decaying learning curve, which allows a qualitative understanding of learning curves (see \cite{Bordelon_Canatar_Pehlevan_2020,Canatar_Pehlevan_2022,Wei_Hu_Steinhardt_2022} for numerical studies and \cite{Tomasini_Sclocchi_Wyart_2022} for a critique of the theory).

\begin{figure}[htb!]
\begin{center}
\centerline{\includegraphics[width=9cm]{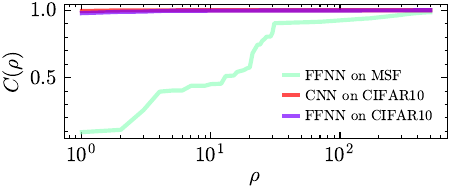}}
\vskip -0.1in
\caption{Cumulative power distribution for a FFNN trained on merged staircase functions as well as a FFNN and CNN trained on CIFAR-10. This can correctly predict the different generalization errors observed in \cref{fig_fig1}.}
\label{fig_cumm}
\end{center}
\vskip -0.2in
\end{figure}
\FloatBarrier


\section{Related work}
\label{appx_relwork}
For a discussion of FL definitions, we refer readers to \cref{sec_current_fl_def}, which provides a thorough literature review. See \cref{tab:nn_kernel_comparison} for an overview of datasets where kernels and NNs provably show a separation in sample complexity.
In the line of works on sample complexity, we highlight several studies on NN scaling laws \cite{Bordelon_Atanasov_Pehlevan_2024,Malach_Kamath_Abbe_Srebro_2021,Shi_Wei_Liang_2022}, alongside theoretical predictions of learning curves in the infinite width limit in \cite{Cohen_Malka_Ringel_2021}. Recent work has critically examined the explanatory power of the NTK for NN generalization \cite{Ortiz-Jiménez_Moosavi-Dezfooli_Frossard_2021,Vyas_Bansal_Nakkiran_2022,Wenger_Dangel_Kristiadi_2024,Ghorbani_Mei_Misiakiewicz_Montanari_2020}.
For a statistical physics-inspired predictive FL theory, we refer to \cite{Aiudi_Pacelli_Baglioni_Vezzani_Burioni_Rotondo_2025} and \cite{Seroussi_Naveh_Ringel_2022}.

\section{Data sets and experiment details}
\subsection{Functions with the merged-staircase property}
\label{appx_sec_msp_f}
In this section we  follow \cite{Abbe_Boix-Adsera_Misiakiewicz_2024}. For any function $f: \{+1, -1\}^d \rightarrow \mathbb{R}$, we can express it using the Fourier-Walsh basis decomposition
\begin{equation}
f(z) = \sum_{S \subseteq [d]} \hat{f}(S)\chi_S(z), z\in \{+1, -1\}^d,
\end{equation}
with Fourier coefficients $\hat{f}(S)$ and basis functions $\chi_S(z) := \prod_{i\in S} z_i$.  This provides a representation of $f(z)$ through orthogonal monomials $\chi_S(z)$ weighted by their respective Fourier coefficients.
\begin{definition}[Merged-Staircase Property]
We say a set structure $\mathcal{S} = \{S_1, \ldots, S_m\} \subseteq 2^{[d]}$ exhibits the Merged-Staircase Property (MSP) if there exists an ordering where each set $S_i$, $i \in [m]$, satisfies:
\begin{equation}
|S_i \setminus \cup_{i' < i} S_{i'}| \geq 1.
\end{equation}
\end{definition}
This property ensures that each set contributes at least one novel element not contained in the union of preceding sets.
\begin{definition}[Merged-Staircase Property for Functions]
Let $\mathcal{S} \subset 2^{[d]}$ be non-zero Fourier coefficients of $f$. We say that $f$ satisfies the Merged-Staircase Property (MSP) if $\mathcal{S}$ has a MSP set structure.
\end{definition}
In the empirical experiments, we use $f(z) = z_7 + z_2z_7 + z_0z_2z_7 + z_4z_5z_7 + z_1 + z_0z_4 + z_3z_7 + z_0z_1z_2z_3z_4z_6z_7$ with $d=30$.

\subsection{Multi-index functions}
\label{appx_sec_pp_f}
Here, we follow \cite{Damian_Lee_Soltanolkotabi_2022}. Multi-index functions are polynomials  that depend on a small number of latent directions. Following Assumptions 1 and 2 in \cite{Damian_Lee_Soltanolkotabi_2022} from the theoretical analysis, we construct functions of the form $f(\vx) = g(\langle \vx, \bm{u}_1 \rangle, \ldots, \langle \vx, \bm{u}_r \rangle)$ where $\{\bm{u}_1, \ldots, \bm{u}_r\}$ are linearly independent vectors spanning the principal subspace $S^*$, while ensuring the non-degeneracy condition that the expected Hessian $H = \mathbb{E}_{\vx\sim\mathcal{D}}[\nabla^2f(\vx)]$ has rank exactly $r$.

Specifically, we first construct a random orthogonal projection matrix $U \in \mathbb{R}^{d \times r}$ through QR decomposition of a Gaussian random matrix, where $r \ll d$ represents the intrinsic dimension of the target function. This ensures linear independence of the latent directions. Input data is sampled from a standard normal distribution $X \sim \mathcal{N}(0, I_d)$ and projected onto this latent space via $X_{latent} = XU$. The target polynomial function is then constructed as a sum over all multi-indices $\alpha \in \mathbb{N}^r$ with total degree at most $p$, where each term has a random Gaussian coefficient $c_\alpha \sim \mathcal{N}(0,1)$: $f(\vx) = \sum_{\alpha: \|\alpha\|_1 \leq p} c_\alpha \prod_{i=1}^r (U^T \vx)_i^{\alpha_i}$. 

To model measurement noise, we add symmetric binary noise $\epsilon \sim \mathcal{U}(\{-\sigma, \sigma\})$ to obtain the final labels $y = f(\vx) + \epsilon$. For evaluation, we generate test data using the same projection matrix $U$ and coefficients $c_\alpha$ but with fresh input samples. The outputs are normalized to have zero mean and unit variance based on training set statistics.

\section{$\mu$P-parameterization}
When training with $\mu$P parameterization, we follow the definition of the $\mu$P parameterization in \cite{Everett_Xiao_Wortsman_Alemi_Novak_Liu_Gur_Sohl-Dickstein_Kaelbling_Lee_et}.
For an $L$-layer feed-forward NN with width $n_l$ and input dimension $n_0$, muP prescribes specific initialization and learning rate scaling rules:
\paragraph{Initialization}
The weights $\vW^l$ at each layer are initialized as:
\begin{equation}
\begin{aligned}
\vW^1 &\sim \mathcal{N}\left(0, \frac{1}{n_0}\right), \quad 
\vW^l &\sim \mathcal{N}\left(0, \frac{1}{n_{l-1}}\right) \quad  2 \leq l \leq L, \quad 
\vW^{L+1} &\sim \mathcal{N}\left(0, \frac{1}{n_L}\right) 
\end{aligned}
\end{equation}
All bias terms are initialized to zero:
\begin{equation}
\vb^l = \mathbf{0} \quad \forall l \in {1,\ldots,L+1}
\end{equation}
\paragraph{Learning Rate Scaling}
The learning rates $\eta_l$ for each layer follow:
\begin{equation}
\begin{aligned}
\eta_1 &= \eta_{\text{base}}, \quad
\eta_l &= \frac{\eta_{\text{base}}}{n_{l-1}} \quad 2 \leq l \leq L ,\quad
\eta_{L+1} &= \frac{\eta_{\text{base}}}{n_L} 
\end{aligned}
\end{equation}
where $\eta_{\text{base}}$ is the base learning rate.
\subsection{Figure 1}

\begin{table}[htb!]
\centering
\begin{tabular}{|l|c|c|}
\hline
\textbf{Hyperparameter} & \textbf{MSP Experiment} & \textbf{Multi-index functions Experiment} \\
\hline

latent dimension & --- & 3 \\
polynomial degree & --- & 5 \\
noise std & --- & 0.0 \\
$P$ (MSP parameter) & 8 & --- \\
$d$ (input dimension) & 30 & 20 \\
\# hidden layer & 4 &  4 \\
hidden layer sizes & [400] & [400] \\
activation & ReLU & ReLU \\
batch size & 64 & 64 \\
epochs & 5000 & 5000 \\
learning rate & 0.05 & 0.001 \\
weight decay & $10^{-4}$ & $10^{-4}$ \\
initialization mode & muP Pennington & muP Pennington \\
$\gamma$ & 1 & 1\\
test set size & 1000 & 10000 \\
optimizer & Adam (muP mode) & Adam (muP mode) \\
learning rate scheduler & CosineAnnealing & CosineAnnealing \\
gradient clipping & 1.0 & 1.0 \\

MSP sets & \multicolumn{2}{l|}{$\{7\}, \{2,7\}, \{0,2,7\}, \{5,7,4\}, \{1\}, \{0,4\},$} \\
& \multicolumn{2}{l|}{$\{3,7\}, \{0,1,2,3,4,6,7\}$} \\
\hline
\end{tabular}
\caption{ Hyperparameter settings for both MSP and multi-index functions experiments in \cref{fig_fig1} (a), (b).}
\label{tab:hyperparameters-combined}
\end{table}

\FloatBarrier

\begin{table}[htb!]
\centering
\begin{tabular}{|l|c|}
\hline
\textbf{Hyperparameter} & \textbf{Value} \\
\hline
Architecture & WideResNet \cite{Zagoruyko_Komodakis_2017} \\
Block size & 4 \\
Width multipliers ($k$) & 4.0 \\
Number of classes & 10 \\
Initial channels & 16 \\
Channel progression & [16$k$, 32$k$, 64$k$] \\
Dataset & CIFAR-10 \\
Input normalization & divide by 255 \\
Batch size & 128 \\
Epochs & 200 \\
Learning rate & 0.001 \\
Optimizer & Adam \\
Loss function & MSE with one-hot targets \\
Learning rate scheduler & CosineAnnealing \\
Number of test samples & 10000 \\
\hline
\end{tabular}
\caption{ Hyperparameter settings for training on CIFAR-10  \cref{fig_fig1} (c).}
\label{tab:hyperparameters-wideresnet}
\end{table}
\FloatBarrier

\subsection{Figure 2}

\begin{table}[htb!]
\centering
\begin{tabular}{|l|c|}
\hline
\textbf{Hyperparameter} & \textbf{Value} \\
\hline
Architecture & FFNN \\
Hidden sizes & [400, 1000] \\
Depth & 4 \\
Weight initialization & He ($1/\sqrt{N}$) \\
Input dimension ($d$) & 30 \\
\hline
\multicolumn{2}{|l|}{\textbf{Training Parameters}} \\
\hline
Test set size & 5000 \\
Training set sizes & [10, 100, 250, 500, 750, 1000,\\
& 2500, 5000, 7500, 10000, 20000] \\
Batch size & 64 \\
Epochs & 3000 \\
Learning rate & 0.005 \\
Weight decay & $10^{-4}$ \\
Optimizer & AdamW \\
LR scheduler & Cosine decay \\
Gradient clipping & 1.0 \\
$\gamma$ scaling & [1.0,0.01] \\
Number of experiments & 3 \\

\hline
\multicolumn{2}{|l|}{\textbf{MSP Parameters}} \\
\hline
$d$  & 30 \\
MSP sets & $\{7\}, \{2,7\}, \{0,2,7\}, \{5,7,4\},$\\
& $\{1\}, \{0,4\}, \{3,7\}, \{0,1,2,3,4,6,7\}$ \\
\hline
\multicolumn{2}{|l|}{\textbf{NTK Parameters}} \\
\hline
NTK computation & Empirical, batched \\
Kernel regularization & $10^{-6} \cdot \text{tr}(K)/n$ \\

\hline
\end{tabular}
\caption{Hyperparameter settings for the experiments with FFNNs in \cref{fig_cifar_shuffle}.}
\label{tab_fig2_FFNN}
\end{table}

\FloatBarrier

\begin{table}[htb!]
\centering
\begin{tabular}{|l|c|}
\hline
\textbf{Hyperparameter} & \textbf{Value} \\
\hline
Architecture & WideResNet \cite{Zagoruyko_Komodakis_2017} \\
Block size & 4 \\
Width multiplier ($k$) & 2.0 \\
Number of classes & 10 \\
Initial channels & 16 \\
Channel progression & [16$k$, 32$k$, 64$k$] \\
Normalization & LayerNorm \\
\hline
\multicolumn{2}{|l|}{\textbf{Training Parameters}} \\
\hline
Input normalization & $\frac{x - \mu}{\sigma}$ (per channel) \\
Batch size & 64 \\
Epochs & 2500 \\
Base learning rate & 0.0001 \\
Weight decay & $10^{-4}$ \\
Optimizer & AdamW \\
Loss function & Cross-entropy \\
LR scheduler & Cosine decay \\
Gradient clipping & 10.0 \\
Training set sizes & [10, 100, 500, 1000, 2000,\\
& 4000, 8192, 16384, 32768] \\
Test set size & 10000 \\
Number of experiments & 3 \\
\hline
\multicolumn{2}{|l|}{\textbf{NTK Parameters}} \\
\hline
NTK computation & Empirical, batched \\
Kernel regularization & $10^{-2}$ if $n > 8000$ else $10^{-4}$ \\
\hline
\end{tabular}
\caption{Hyperparameter settings for experiments with the WideResNet  in \cref{fig_cifar_shuffle}.}
\label{tab_fig2_CNN}
\end{table}

\FloatBarrier

\subsection{\cref{fig_yoonsoo_quality,fig_anthro_main}}
\cref{fig_yoonsoo_quality} uses the settings from \cref{fig_fig1}.

\begin{table}[htb!]
\centering
\begin{tabular}{|l|c|}
\hline
\textbf{Hyperparameter} & \textbf{MSP Experiment} \\
\hline
$P$ (MSP parameter) & 8 \\
$d$ (input dimension) & 30 \\
\# hidden layer & 4 \\
hidden layer sizes & 2000 \\
activation & ReLU \\
batch size & 64 \\
epochs & 5000 \\
learning rate & 0.001 \\
weight decay & $10^{-4}$ \\
initialization mode & muP Pennington \\
$\gamma$ & [1, 0.0001] \\
test set size & 1000 \\
optimizer & Adam (muP mode) \\
learning rate scheduler & CosineAnnealing \\
gradient clipping & 1.0 \\
MSP sets & $\{7\}, \{2,7\}, \{0,2,7\}, \{5,7,4\},$\\
& $\{1\}, \{0,4\}, \{3,7\}, \{0,1,2,3,4,6,7\}$ \\
\hline
\end{tabular}
\caption{Hyperparameter settings for for \cref{fig_anthro_main}.}
\label{tab_hyperfig45}
\end{table}

\FloatBarrier

\subsection{Addition to: \cref{sec_flgap}}

\begin{table}[htb!]
\setlength{\tabcolsep}{4pt}
\footnotesize
\begin{tabular}{lccccc}
\toprule
\textbf{Source} & \textbf{Data} & \textbf{NN Type} & \textbf{Kernel} & \textbf{NN Scaling} & \textbf{Kernel Scaling} \\
\midrule
\cite{Damian_Lee_Soltanolkotabi_2022}, & $x \sim \mathcal{N}(0,I_d)$, & 1-hidden layer & NTK & $m =\Omega(dr^p +$ & $m = \Omega(\frac{d^{p/2}}{\varepsilon^2})$ \\
\cite{Donhauser_Wu_Yang_2021}, & $y = g(Ux)$, & $N=O(r^p)$ & & $\frac{d^2r}{\varepsilon^2})$ & \\
\cite{Ghorbani_Mei_Misiakiewicz_Montanari_2020}, & $U \in \mathbb{R}^{r \times d}$, & & & & \\
\cite{Mousavi-Hosseini_Park_Girotti_Mitliagkas_Erdogdu_2023}, & deg $p$ poly & & & & \\
\cite{Troiani_Dandi_Defilippis_Zdeborová_Loureiro_Krzakala_2024}\\
\cite{,Mousavi}\\
\midrule
\cite{Abbe_Boix-Adsera_Misiakiewicz_2024} & MSP function, input dim. $d$ &1-hidden layer NN & Any & $m = O(d \cdot 2^{2^{O(P)}}/\varepsilon^5)$ & $n = \Omega(d^P)$ \\
& max. poly. degree. $P$ &   with $N=e^{e^{P}}$& &  &  \\
\midrule
\cite{Daniely_Malach_2020}, & Sparse parity & 1-hidden layer & Any & $m = \Omega(\text{poly}(k,$ & $m = \Omega(\frac{2^k}{\varepsilon^2})$ \\
\cite{Telgarsky} & on $k$ bits & $N=\text{poly}(k)$ & & $\frac{1}{\varepsilon}))$ & \\
\midrule
\cite{Refinetti_Goldt_Krzakala_Zdeborová_2021} & $d$-dim Gaussian mixture & 1-hidden layer & ReLU RFK & $\epsilon_{NN} = \Theta(1)$ & $\epsilon_{RF} = \Omega(1)$ \\
& 4 clusters XOR config & $O(1)$ width & $O(d)$ feats & $m=\Omega(d)$ & $m=O(d)$ \\
\midrule
\cite{Frei_Chatterji_Bartlett}, & Noisy 2-XOR cluster & 1-hidden layer & NTK & $m=O(\frac{1}{\varepsilon^2})$ & $m=\Omega(\frac{d^2}{\varepsilon^2})$ \\
\cite{Wei_Lee_Liu_Ma_2019} & $d$-dim distribution & & & & \\
\midrule
\cite{Székely_Bardone_Gerace_Goldt_2024} & Spiked $d$-dim cumulant & 1-hidden layer & ReLU RFK & $\epsilon_{NN} = O(1)$ & $\epsilon_{RF} = \frac{1}{2}$ \\
& model ($\geq 4$ cumulants) & $N\geq 5d$ & $O(d)$ & $m = \Omega(d^2)$ & $m = O(d^2)$ \\
\midrule
\cite{Akiyama_Suzuki_2022}, & Uniform on $S^{d-1}$, & 1-hidden layer & Any & $m=O(\frac{N^5}{\varepsilon})$ & $m=\Omega({\varepsilon}^{-\frac{d+2}{2d+2}})$ \\
\cite{Allen-Zhu_Li_2020}, & Two-layer ReLU & width $N$ & & fixed width & \\
\cite{Allen-Zhu_Li_2023}, & teacher network & & & & \\
\cite{Li_Ma_Zhang_2020}, & width $N$ & & & & \\
\cite{Yehudai_Shamir_2022} & & & & & \\
\bottomrule
\end{tabular}
\caption{Comparison of NN and kernel method scaling for different target functions.}
\label{tab:nn_kernel_comparison}
\end{table}
\FloatBarrier

\subsection{Addition to: \cref{sec_current_fl_def}}
\subsection{2. Family: Spectra}

\label{appx_sec_2family_spectra}
\begin{figure}[htb!]
\begin{center}
\centerline{\includegraphics[width=12cm]{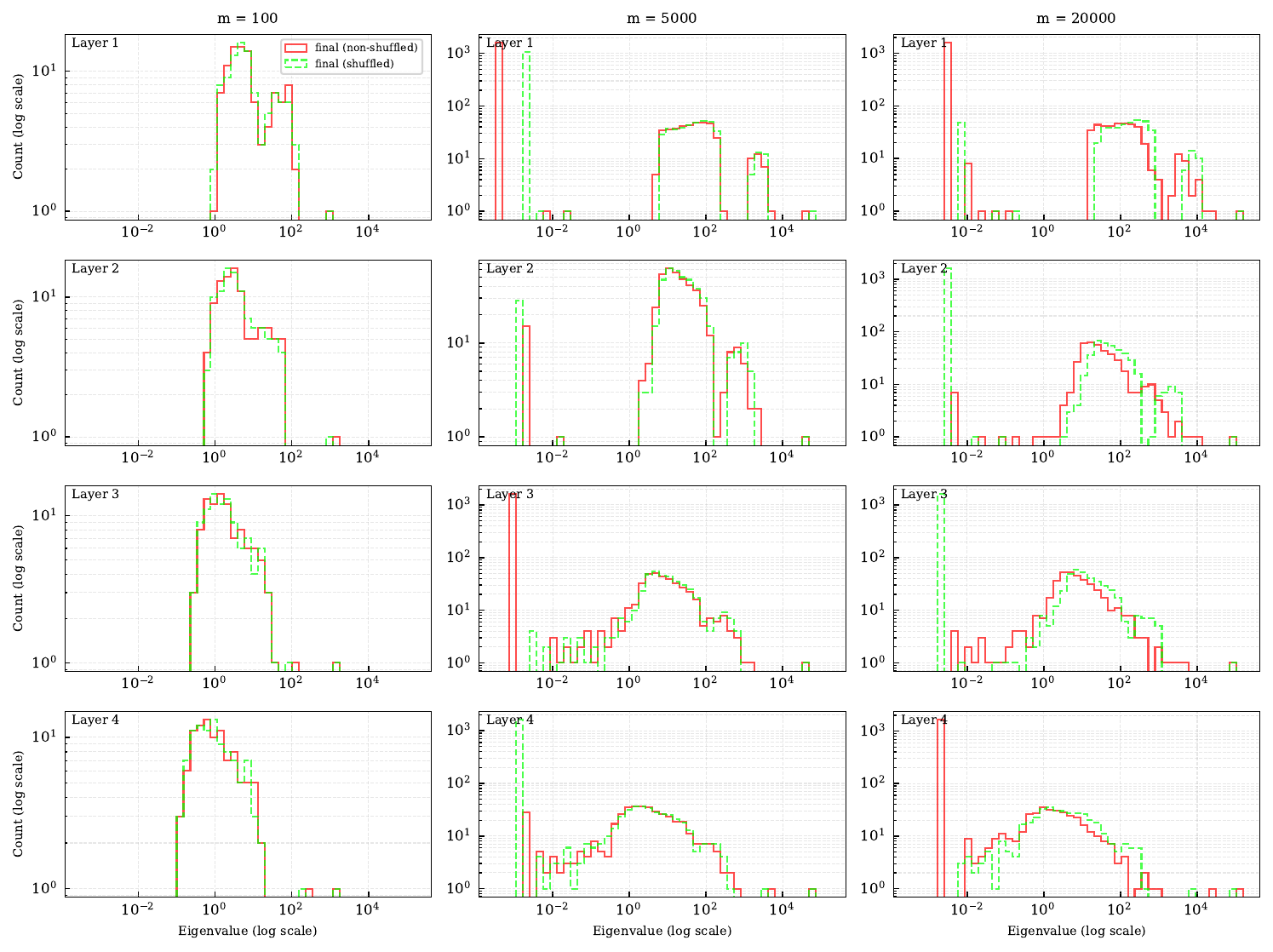}}
\vskip -0.1in
\caption{CK spectrum for a NN trained with standard parameterization and $N=400$. }
\label{appx_cksp1}
\end{center}
\vskip -0.2in
\end{figure}
\FloatBarrier

\begin{figure}[htb!]
\begin{center}
\centerline{\includegraphics[width=12cm]{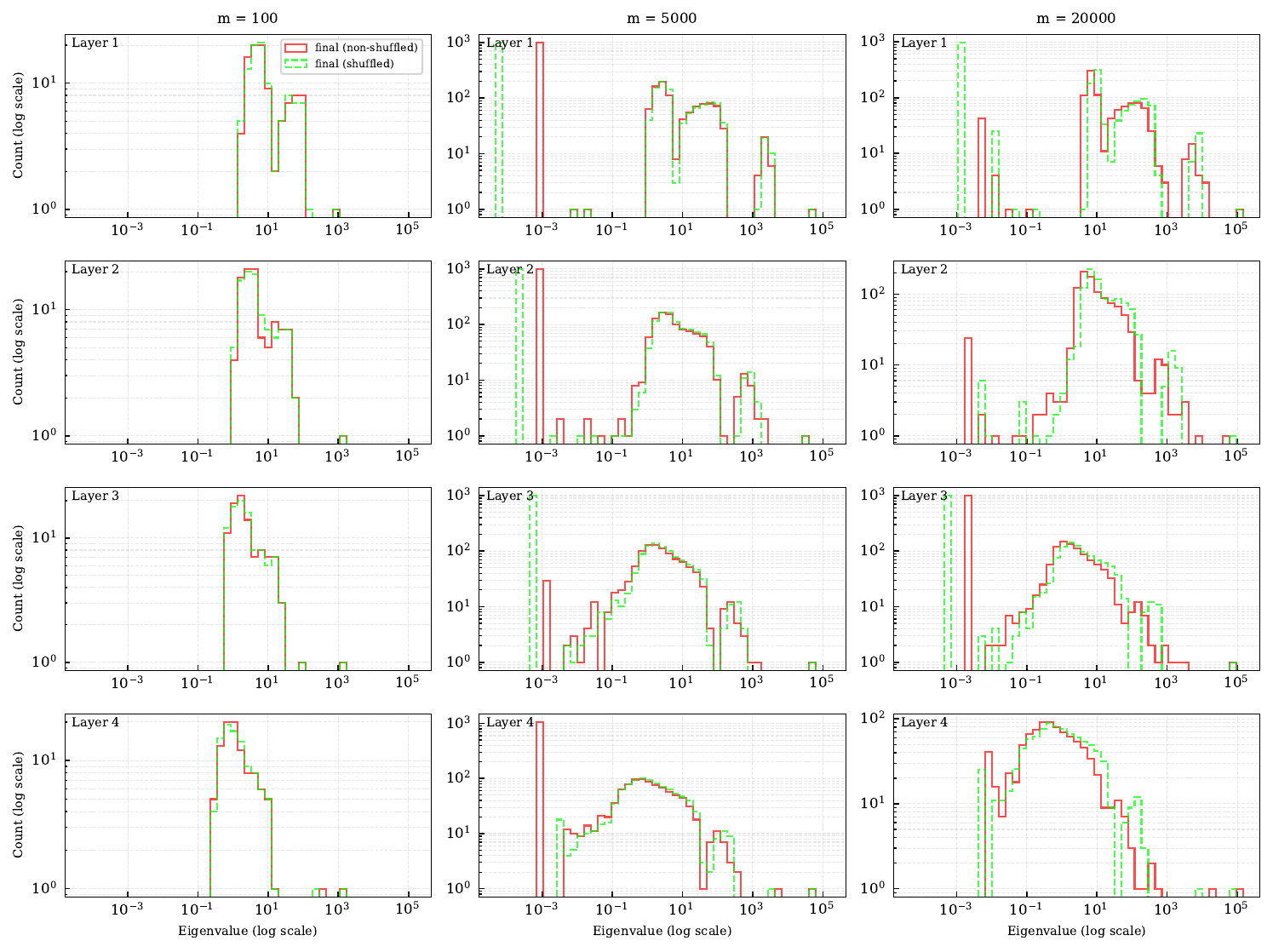}}
\vskip -0.1in
\caption{CK spectrum for a NN trained with standard parameterization and $N=1000$. }
\label{}
\end{center}
\vskip -0.2in
\end{figure}
\FloatBarrier

\begin{figure}[htb!]
\begin{center}
\centerline{\includegraphics[width=12cm]{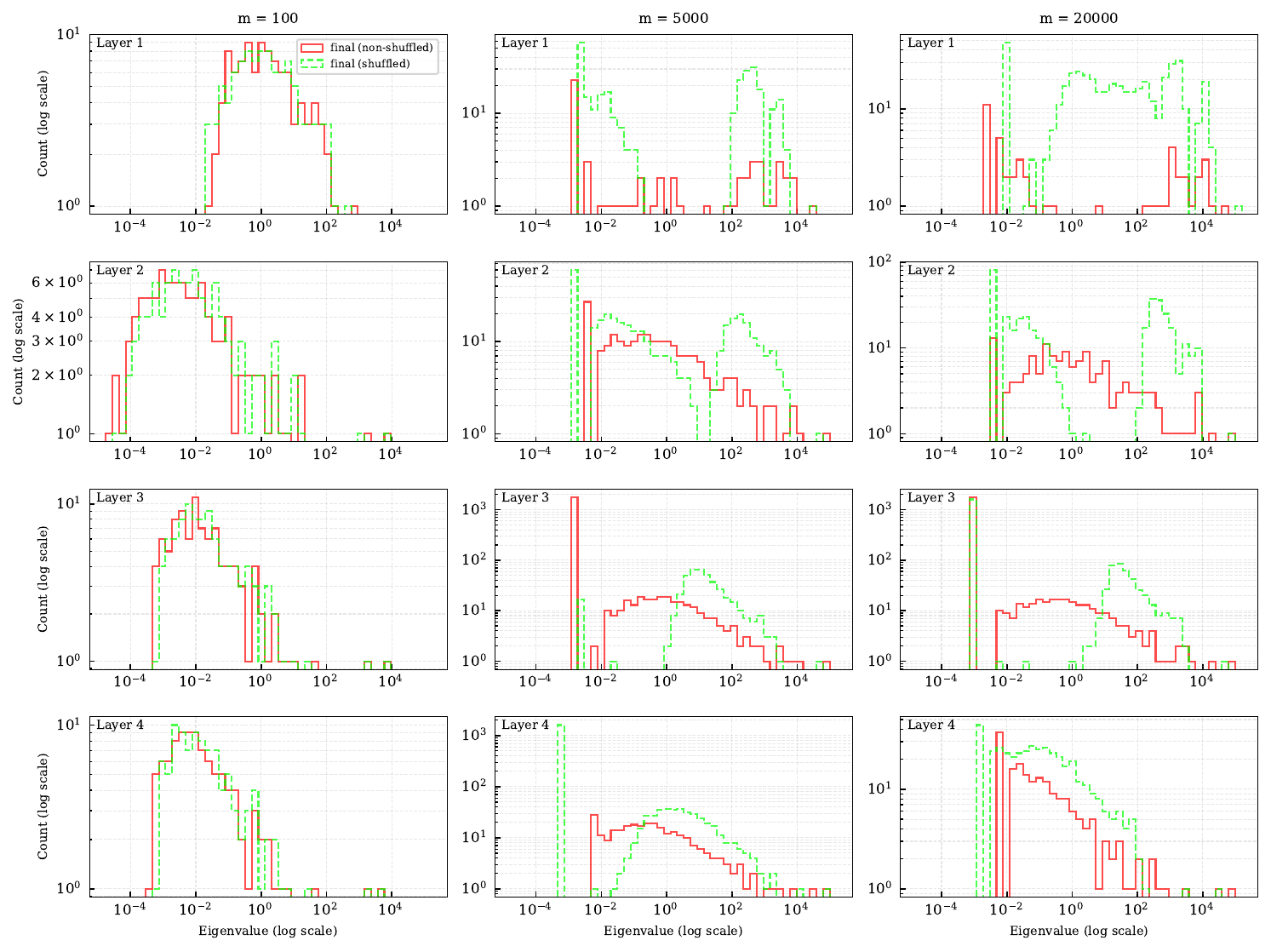}}
\vskip -0.1in
\caption{CK spectrum for a NN trained with  $\mu$P-parameterization and $N=400$. }
\label{}
\end{center}
\vskip -0.2in
\end{figure}
\FloatBarrier

\begin{figure}[htb!]
\begin{center}
\centerline{\includegraphics[width=12cm]{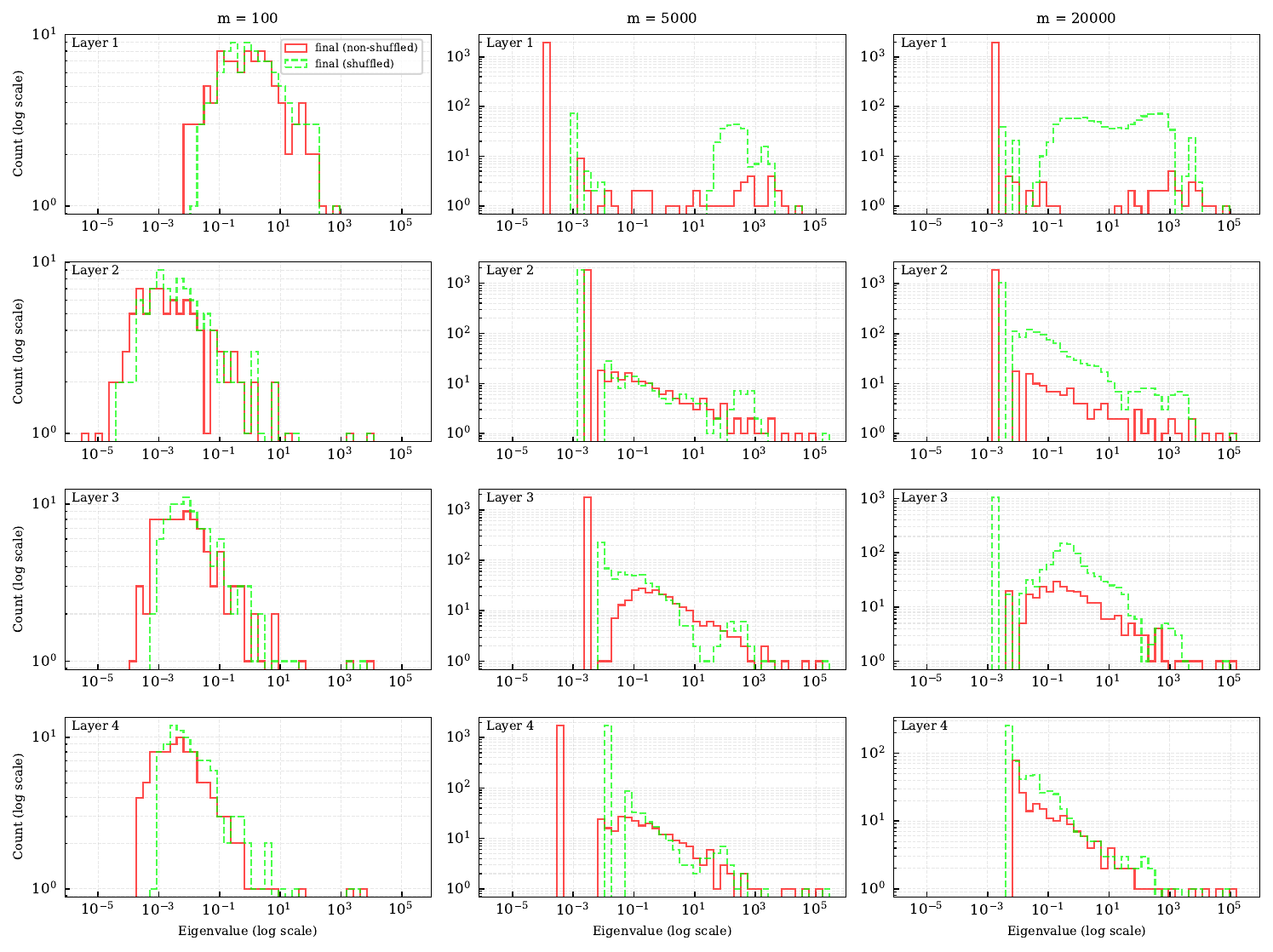}}
\vskip -0.1in
\caption{CK spectrum for a NN trained with $\mu$P-parameterization and $N=1000$. }
\label{appx_cksp4}
\end{center}
\vskip -0.2in
\end{figure}
\FloatBarrier

\subsection{3. Family: Distribution plots}
\begin{figure}[htb!]
\begin{center}
\centerline{\includegraphics[width=15cm]{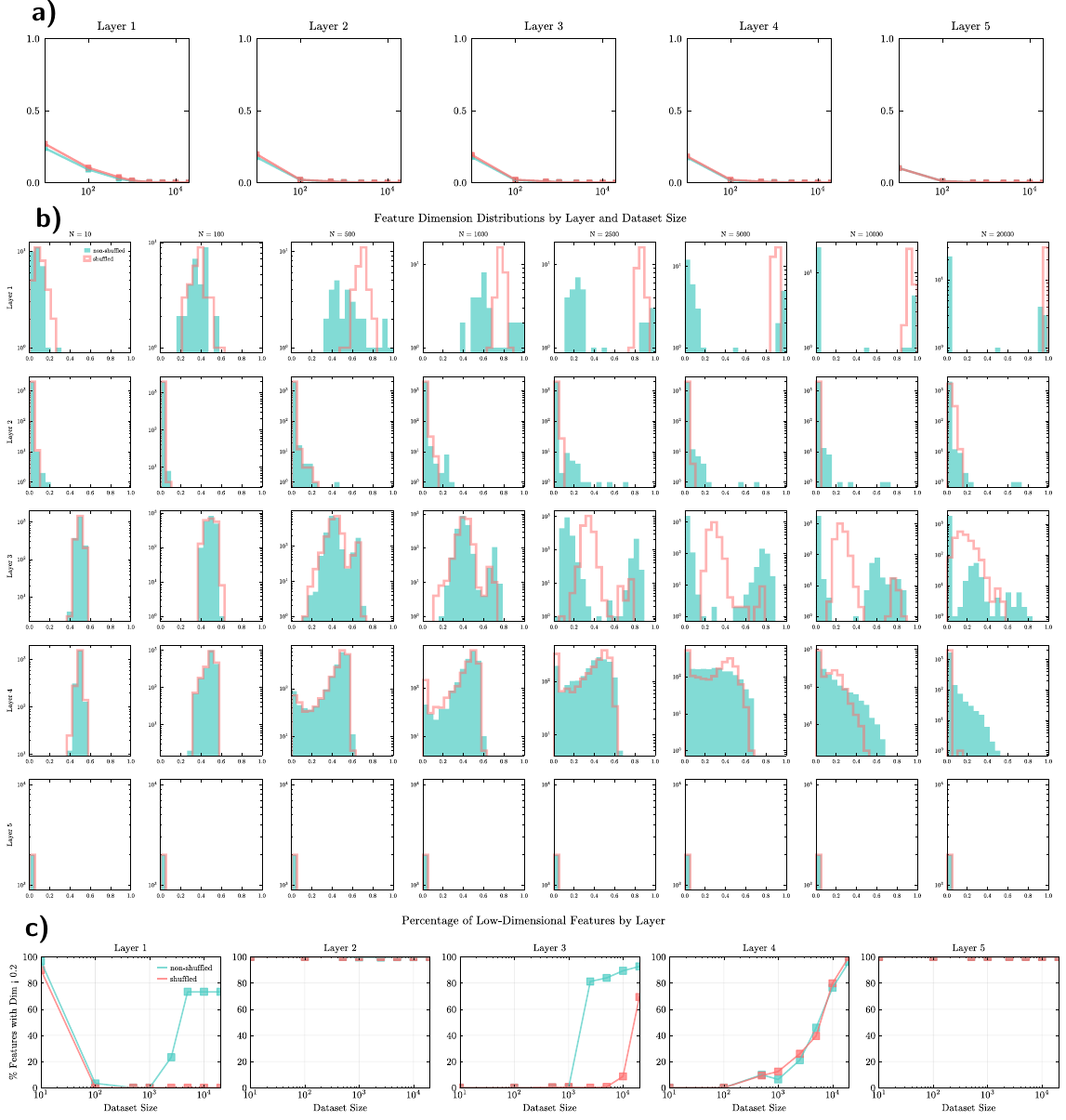}}
\vskip -0.1in
\caption{NNs (width 2000) trained with $\mu$P parameterization on MSP functions across varying training set sizes $m$ with $\gamma=1$. (a) Sample complexity analysis fails to differentiate between shuffled and non-shuffled data. (b) Per-layer feature dimensionality comparison between shuffled and non-shuffled datasets reveals diffuse patterns across training set sizes. (c) Relative number of dimensions with $D_{f_i}=0$. The proportion for the first layer rises around $m^*$. This is not observed in later layers.}
\label{fig_anthrohistg1}
\end{center}
\vskip -0.2in
\end{figure}
\FloatBarrier

\begin{figure}[htb!]
\begin{center}
\centerline{\includegraphics[width=15cm]{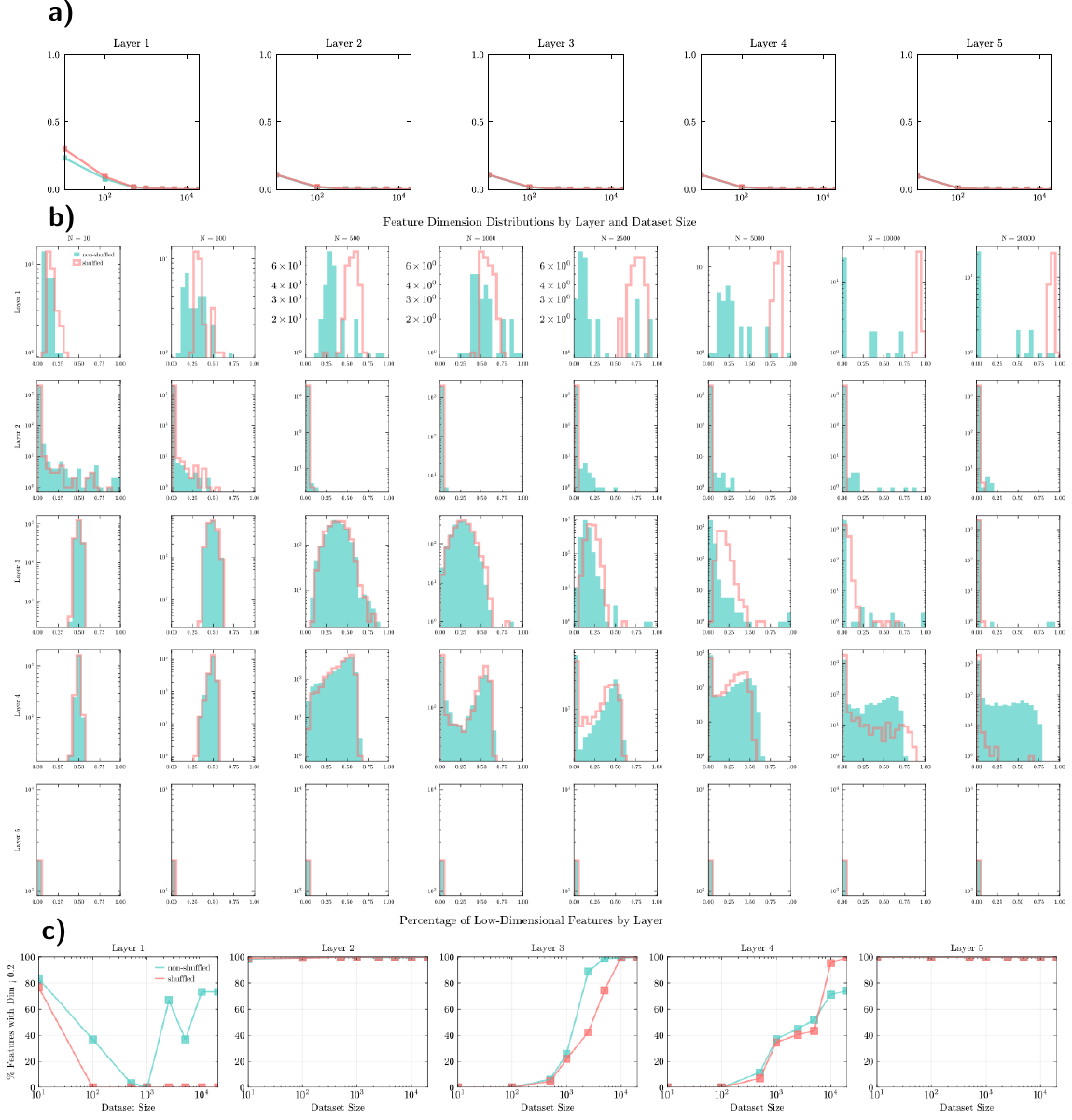}}
\vskip -0.1in
\caption{Same as \cref{fig_anthrohistg1} but with $\gamma=0.0001$. $\gamma=0.0001$ moves the weight of the $D_{f_i}$ distributions closer to 0.}
\label{fig_anthrohistg00001}
\end{center}
\vskip -0.2in
\end{figure}
\FloatBarrier




\subsection{Additional information on $\Delta_{\operatorname{NT}}$ }
\label{appx_sec_conc}

\begin{figure}[htb!]
\vskip 0.2in
\begin{center}
\centerline{\includegraphics[width=14cm]{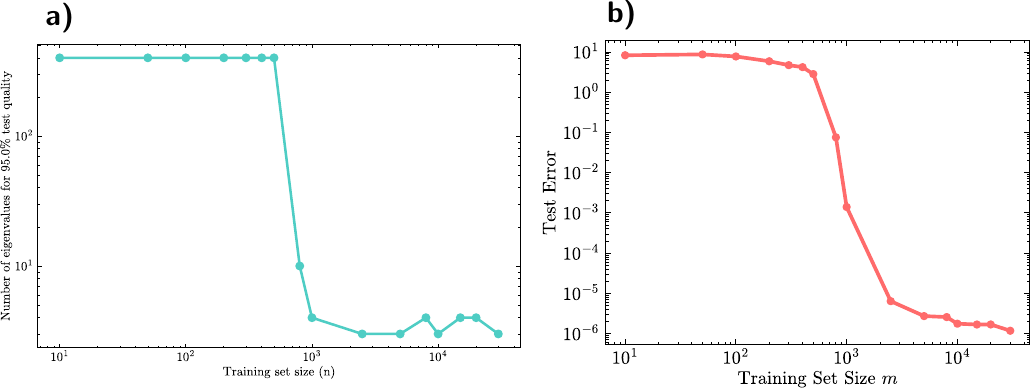}}
\caption{ NNs (width 400, depth 4) trained with $\mu$P on MSP functions. (a) $S_{\operatorname{CK}}$ computed via projection onto the target function using $Q_k=\braket{e_k|f^*}$ instead of the learned function, quantifying how well the top $k$ eigenfunctions approximate the target function. (b) Generalization error versus training set size. This  demonstrates that $S_{\operatorname{CK}}$  can predict the generalization error as both curves strongly correlate.  }
\label{fig_corr2}
\end{center}
\vskip -0.2in
\end{figure}
\FloatBarrier

In the following we will define the critical dataset size $m^*$ more formally.
\begin{definition}
   Given a data generating model $p(\vx,\vy)$ and an i.i.d. dataset $\mathcal{D}$ of size $m$, a FFNN $f_{\theta}$ of width $N$ and depth $D$ with $\mu$P-parameterization and base learning rate $\eta_0$, we define the critical training set size $m^*$ as the smallest dataset size such that
    \begin{equation}
       \exists N^*: \quad  \forall N \geq N^*,   \forall m \geq m^*: \quad \frac{\mathcal{E}(f^*_{\theta};m)}{\mathcal{E}(f^*_{NTK};m)}< \varepsilon
    \end{equation}
    where the generalization error of the NN is smaller than the one of the NTK by a factor of $\varepsilon$, typically taken to be $\varepsilon \approx 1/10$ or smaller. 
\end{definition}

\paragraph{Dependence on $N,D$} For the $\mu$P-parameterization, we find that $m^*$ exhibits weak dependence on the initial learning rate $\eta_0$ and  depth $D$, while remaining independent of width $N$. Base learning rates that are too small lead to very slow training. As depth increases, the error $\mathcal{E}(f^*_{\theta};m)$ decreases for fixed $m$, generally resulting in smaller values of $m^*$. In contrast, under standard parameterization, $m^*$ shows width dependence—an artifact of suboptimal hyperparameter selection rather than an intrinsic property.

\begin{figure}[htb!]
\vskip 0.2in
\begin{center}
\centerline{\includegraphics[width=12.5cm]{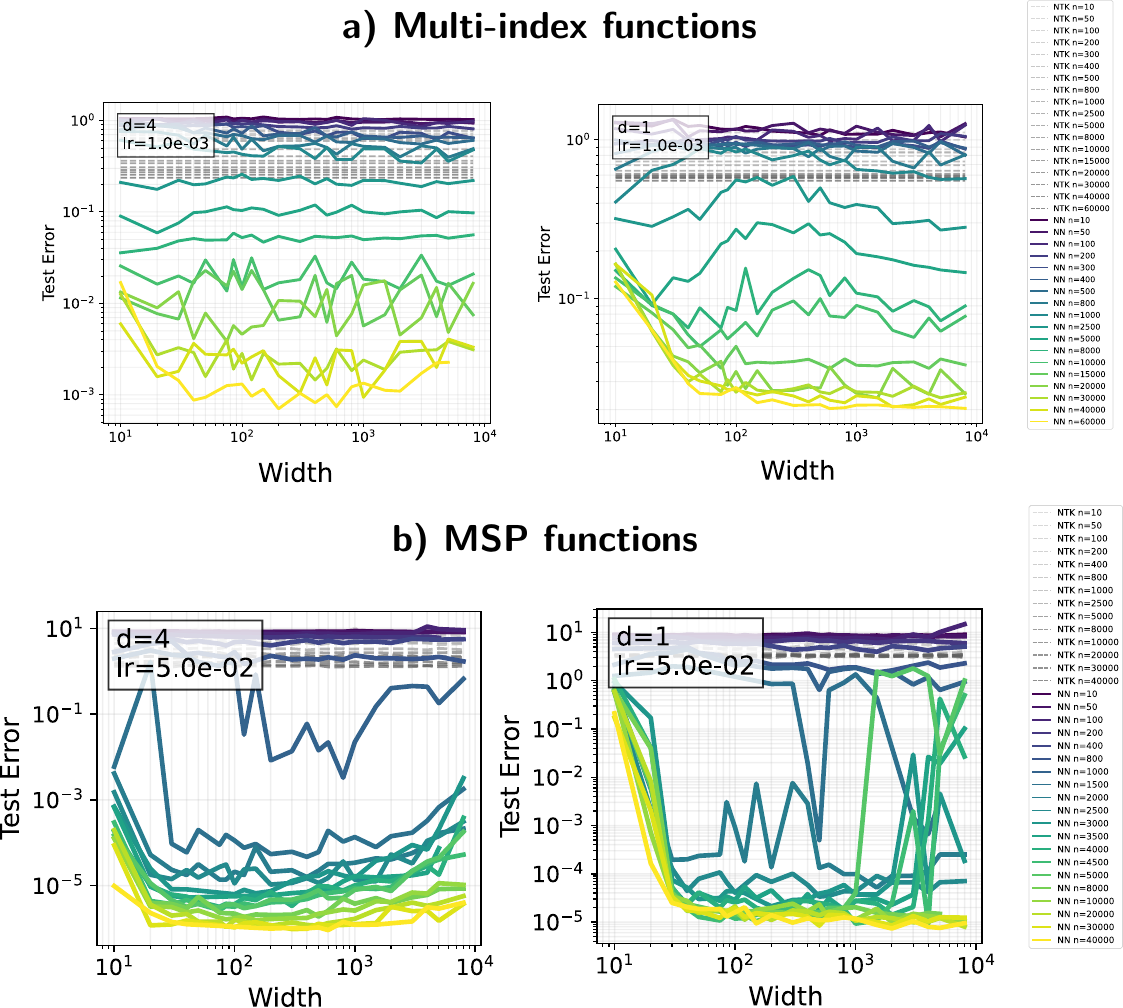}}
\caption{Generalization error of NNs with varying widths and depths ($d=1,4$) trained with $\mu$P on (a) multi-index functions and (b) MSP functions. The results demonstrate a width-independence threshold: beyond a critical width where the network achieves sufficient expressivity, the generalization error remains consistent regardless of further width increases. }
\label{fig_width}
\end{center}
\vskip -0.2in
\end{figure}

\begin{figure}[htb!]
\vskip 0.2in
\begin{center}
\centerline{\includegraphics[width=11cm]{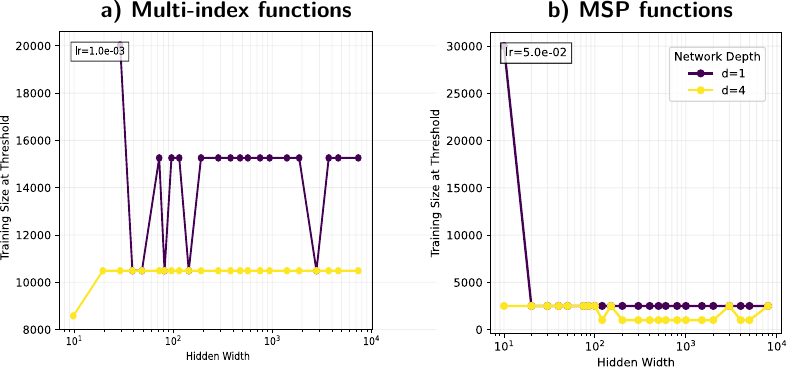}}
\caption{Critical dataset size $m^*$ as a function of network width for (a) multi-index functions and (b) MSP functions, demonstrating that $m^*$ beyond a certain width threshold, exhibits width independence. }
\label{fig_m}
\end{center}
\vskip -0.2in
\end{figure}

\begin{figure}[htb!]
\vskip 0.2in
\begin{center}
\centerline{\includegraphics[width=8cm]{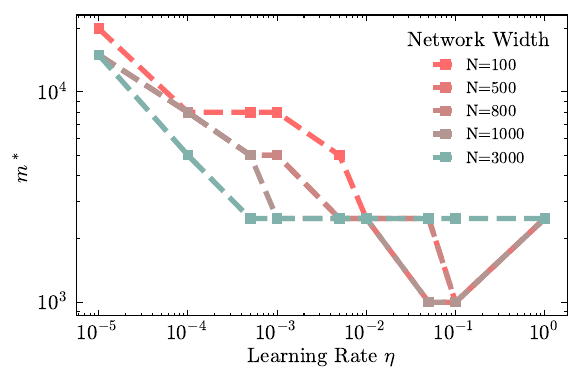}}
\caption{Critical dataset size $m^*$ plotted against base learning rate for MSP functions, revealing a stable region where $m^*$ remains constant across a specific range of learning rates.}
\label{fig_lr}
\end{center}
\vskip -0.2in
\end{figure}

\end{document}